\documentclass{article}
 
\usepackage[english]{babel}
\usepackage[utf8x]{inputenc}  
\usepackage[T1]{fontenc}
\usepackage{stmaryrd} 

\usepackage[a4paper,top=3cm,bottom=2cm,left=3cm,right=3cm,marginparwidth=1.75cm]{geometry}
\usepackage{tikz}
\usetikzlibrary{arrows,calc,fit,arrows.meta}
\usepackage{amsmath}
\usepackage{amsfonts}
\usepackage{subcaption}
\usepackage[disable,colorinlistoftodos]{todonotes}
\usepackage{longtable}

\usepackage{pdfpages}
\usepackage{rotating}

\usepackage{longtable} 
\usepackage{rotating} 
\usepackage{dsfont} 

\usepackage{breakcites}

\usepackage{graphicx}
\usepackage{color, colortbl}
\usepackage{authblk}
\definecolor{blue}{rgb}{0.5,0.5,0.5}
\usepackage[hyperindex,breaklinks,colorlinks=true, allcolors=blue]{hyperref}

\author[1, 2]{Timoth\'ee Lesort} 
\author[1]{Natalia D\'iaz-Rodr\'iguez}
\author[2]{Jean-Fran\c{c}ois Goudou}
\author[1]{David Filliat}
\affil[1]{\footnotesize U2IS, ENSTA ParisTech, Inria FLOWERS team, Universit\'e Paris Saclay, Palaiseau, France.\\

\texttt{\{timothee.lesort, natalia.diaz, david.filliat\}@ensta-paristech.fr}}

\affil[2]{\footnotesize Vision Lab, Thales, Theresis, Palaiseau, France.\\

 \texttt{jean-francois.goudou@thalesgroup.com}}

\title{State Representation Learning for Control: An Overview}

\providecommand{\Keywords}[1]{\textbf{\textit{Keywords: }} #1}
\begin{document}

\date{}
\maketitle

\begin{abstract} 

Representation learning algorithms are designed to learn abstract features that characterize data.
State representation learning (SRL) focuses on a particular kind of representation learning where learned features are in low dimension, evolve through time, and are influenced by actions of an agent.
The representation is learned to capture the variation in the environment generated by the agent's actions; this kind of representation is particularly suitable for robotics and control scenarios.
In particular, the low dimension characteristic of the representation helps to overcome the curse of dimensionality, provides easier interpretation and utilization by humans and can help improve performance and speed in policy learning algorithms such as reinforcement learning.

This survey aims at covering the state-of-the-art on state representation learning in the most recent years. It reviews different SRL methods that involve interaction with the environment, their implementations 
and their applications in robotics control tasks (simulated or real). In particular, it highlights how generic learning objectives are differently exploited in the reviewed algorithms. Finally, it discusses evaluation methods to assess the representation learned and summarizes current and future lines of research.

\end{abstract}

\Keywords{State Representation Learning, Low Dimensional Embedding Learning, Learning Disentangled Representations, Disentanglement of control factors, Robotics, Reinforcement Learning}
\newpage

\section{Introduction}
Robotics control and artificial intelligence (AI) in a broader perspective heavily rely on the availability of compact and expressive representations of the sensor data. Designing such representations has long been performed manually by the designer, but deep learning now provides a general framework to learn such representations from data. This is particularly interesting for robotics where multiple sensors (such as cameras) can provide very high dimensional data, while the robot objective can often be expressed in a much lower dimensional space (such as the 3D position of an object in a manipulation task). This low dimensional representation, frequently called the \emph{state} of the system, has the crucial role of encoding essential information (for a given task) while discarding the many irrelevant aspects of the original data. By \textit{Low dimensional}, we mean that the learned state dimension is significantly smaller than the dimensionality of the observation space.

Such state representation is at the basis of the classical reinforcement learning (RL) framework \cite{Sutton98} in which an agent interacts with its environment by choosing an action as a function of the environment state in order to maximize an expected (discounted) reward. Following this framework, we call \textit{observation} the raw information provided by one or several of the robot sensors, and we call \textit{state} a compact depiction of this observation that retains the information necessary for the robot to choose its actions. 

While deep reinforcement learning algorithms have shown that it is possible to learn controllers directly from observations \cite{Mnih15}, reinforcement learning (or other control algorithms) can take advantage of low dimensional and informative representations, instead of raw data, to solve tasks more efficiently \cite{Munk16}. Such efficiency is critical in robotic applications where experimenting an action is a costly operation. In robotics, as well as in machine learning, finding and defining interesting states (or features) for control tasks usually requires a considerable amount of manual engineering. It is therefore interesting to learn these features with as little supervision as possible. The goal is thus to avoid direct supervision using a \textit{true} state, but instead use information about the actions performed by the agent, their consequences in the observation space, and rewards (even if sparse, and when available). Along with this information, one can also set generic constraints on what a good state representation should be \cite{Jonschkowski15,Lesort17}. 

Feature learning in general is a wide domain which aims at decomposing data into different features that can faithfully characterize it. It has been a particular motivation for deep learning to automatically learn a large range of specific feature detectors in high dimensional problems. State representation learning (SRL) is a particular case of feature learning in which the features to learn are low dimensional, evolve through time, and are influenced by actions or interactions.
SRL is generally framed in a control setup constrained to favor small dimensions to characterize an instance of an environment or an object, often with a semantic meaning that correlates with some physical feature. The physical feature can be, for instance, a position, distance, angle or an orientation.
The objective of SRL is to take advantage of time steps, actions, and optionally rewards, to transform observations into states: 
a vector of a reduced set of the most representative features
that is sufficient for efficient policy learning.
It is also worth distinguishing between feature learning on a process that is only observed, and learning the state representation of a process in which the learning agent possesses embodiment and acts. 
The first one considers learning directly from observations, e.g., pixels, and leaves no room for the agent to act.
The latter gives opportunity to exploit more possibilities to learn better representations by balancing between exploration and exploitation, e.g., active learning, or artificial curiosity \cite{Pere18,Pathak17}.

As stated above, learning in this context should be performed without explicit supervision. In this article we therefore focus on SRL where \textit{learning} does not have the \textit{pattern recognition} regression or classification sense, but rather the sense of the process of model building \cite{Lake16}. Building such models can then exploit a large set of objectives or constraints, possibly taking inspiration from human learning.
As an  example, infants expect inertial objects to follow principles of persistence, continuity, cohesion and solidity before appearance-based elements such as color, texture and perceptual goodness. At the same time, these principles help guide later learnings such as object’ rigidity, softness and liquids properties. Later, adults will reconstruct perceptual scenes using internal representations of the objects and their physically relevant properties (mass, elasticity, friction, gravity, collision, etc.) \cite{Lake16}. In the same way, the SRL literature may make use of knowledge about the physics of the world, interactions and rewards whenever possible as a semi-supervision or self-supervision that aids the challenge of learning state representations without explicit supervision.

Recently, several different approaches have been proposed to learn such state representation. In this review paper, our objective is to present and analyze those different approaches, highlight their commonalities and differences, and to propose further research directions. We extend a previously published review 
\cite{Bohmer15} with the most recent and rapidly evolving literature of the  past years and focus on approaches that learn low dimensional 
Markovian representations without direct supervision, i.e., exploiting sequences of observations, actions, rewards and generic learning objectives. The works selected in this survey mostly evaluate their algorithms in simulations where agents interact with an environment. More marginally, some SRL algorithms are tested on real settings such as robotics tasks, e.g., manipulation or exploration as detailed in Section \ref{sec:EvaluationScenarios}. 

In the remainder of the paper, we first introduce the formal framework and notation, then present the objectives that can be used to learn state representations, and discuss the implementation aspects of these approaches before summarizing some current and future lines of research.

\section{Formalism and definitions}
\label{sec:formalism}

 \subsection{SRL Formalism}

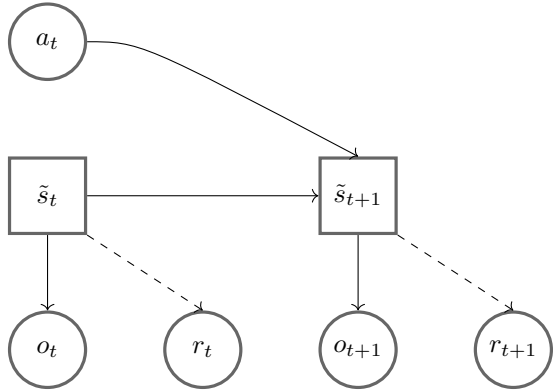
\begin{figure}
\centering
\begin{tikzpicture}[
roundnode/.style={circle, draw=black!60, fill=green!0, very thick, minimum size=10mm},
squarednode/.style={rectangle, draw=black!60, fill=black!0, very thick, minimum size=10mm},
]

\node[squarednode]      (state)                      {$\tilde{s}_t$};
\node[roundnode]        (obs)       [below=of state] {$o_t$};
\node[roundnode]        (reward)       [right=of obs] {$r_t$};
\node[roundnode]        (obs2)       [right=of reward] {$o_{t+1}$};
\node[roundnode]        (reward2)       [right=of obs2] {$r_{t+1}$};
\node[squarednode]      (state2)       [above=of obs2] {$\tilde{s}_{t+1}$};
\node[roundnode]        (action)       [above=of state] {$a_t$};
 
\draw[->] (state2.south) -- (obs2.north);
\draw[->] (state.east) -- (state2.west);
\draw[->] (state.south) -- (obs.north);
\draw[->] (action.east) .. controls +(right:7mm) .. (state2.north);
\draw[dashed, ->] ($(state.east)-(state.north)$) -- (reward.north);
\draw[dashed, ->] ($(state2.east)-(state.north)$) -- (reward2.north);

\end{tikzpicture}

\caption{General model : circle are observable and square are the latent state variables.}
\label{fig:notation}
\end{figure}

The nomenclature we use is very close to the one used in reinforcement learning literature \cite{Sutton98} and is illustrated in Fig. \ref{fig:notation}. We define an environment $\mathcal{E}$ where an agent performs actions $a_t \in \mathcal{A}$ at time step $t$ and where $\mathcal{A}$ is the action space (continuous or discrete). Each action makes the agent transition from a true state $\tilde{s}_t$ to $\tilde{s}_{t+1}$ which is unknown but assumed to exist. We call the true state space $\tilde{\mathcal{S}}$. 
The agent obtains an observation of $\mathcal{E}$ from its sensors, denoted $o_t \in \mathcal{O}$ where $\mathcal{O}$ is the observation space.

Optionally, the agent may receive a reward $r_t$. The reward is given at $\tilde{s}_t$ by a reward function designed to lead the agent to a certain behavior that solves a task. The reward is optional as learning a state representation don't aim to solve a task, but is often present as one of the goal of SRL may be to improve task learning performance.

The SRL task is to learn a representation $s_t \in \mathcal{S}$ of dimension $K$ with characteristics  similar to those of $\tilde{s}_t$ without using $\tilde{s}_t$.
More formally, SRL learns a mapping $\phi$ of the history of observation to the current state $s_t = \phi(o_{1:t})$. Note that actions $a_{1:t}$ and rewards $r_{1:t}$ can also be added to the parameters of $\phi$ \cite{Jonschkowski15}. In this paper, we are specifically interested in the particular setting in which this mapping is learned through proxy objectives without access to the true state $\tilde{s}_t$. 
This family of approaches is called unsupervised or self-supervised.

Finally, we note $\hat{o}_t$ the reconstruction of $o_t$ (similarly for $\hat{a}_t$ and $\hat{r}_t$), that will be used in various SRL approaches.

\subsection{SRL approaches}

Based on the previously defined notations, we can briefly summarize the common strategies used in state representation learning that are detailed in the next sections. In the following, $\theta$ represents the parameters optimized by minimizing the model's loss function. In most of the approaches we present, this model is implemented with a neural network. 

\begin{itemize}

\item{\textbf{Reconstructing the observation}}:
\begin{figure}
\centering
\begin{tikzpicture}[
roundnode/.style={circle, draw=black!60, fill=green!0, very thick, minimum size=10mm},
roundnode2/.style={circle, draw=black!60, fill=black!20, very thick, minimum size=10mm},
squarednode/.style={rectangle, draw=black!60, fill=black!20, very thick, minimum size=10mm},
container/.style={draw, rectangle, draw=blue, dashed, inner sep=1em},
]

\node[squarednode]      (state)                      {$s_t$};
\node[roundnode]        (obs)       [below=of state] {$o_t$};
\node[roundnode2]        (obs2)       [right=of obs] {$ \hat{o}_t$};
\node[container, fit=(obs) (obs2)] (or) {};

\draw[-{Latex[length=3mm,width=2mm]}] (obs.north) -- (state.south);
\draw[-{Latex[length=3mm,width=2mm]}] ($(state.east)-(state.north)$) -- (obs2.north);

\end{tikzpicture}

\caption{Auto-Encoder: reconstructing the observation. The error is computed between observation $o_t$ and its reconstruction $\hat{o}_t$. White components are input data and gray ones are output data.}
\label{fig:AutoEncoder}
\end{figure}
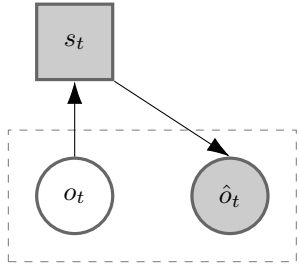
learning the function $\phi$ (Eq. \ref{eq:enc}) so that it is possible to reconstruct the observation with a decoder $\phi^{-1}$ (Eq. \ref{eq:dec}) by minimizing the reconstruction error between the original observation and its predicted reconstruction. The reconstruction is learned under different constraints that give to $s_t$ specific characteristics (e.g., dimensionality constraints, local denoising criterion \cite{Vincent10}, sparse encoding constraints \cite{Vincent08}, etc.) (Fig. \ref{fig:AutoEncoder}).

\begin{equation}  
s_{t} = \phi(o_t; \theta_{\phi})
\label{eq:enc}
\end{equation}

\begin{equation}  
\hat{o}_{t} = \phi^{-1}(s_t; \theta_{\phi^{-1}})
\label{eq:dec}
\end{equation}

where $\theta_{\phi}$ and $\theta_{\phi^{-1}}$ are the parameters learned for the encoder and decoder, respectively.

\item{\textbf{Learning a forward model}}:
\begin{figure}
\centering
\begin{tikzpicture}[
roundnode/.style={circle, draw=black!60, fill=green!0, very thick, minimum size=10mm},
squarednode/.style={rectangle, draw=black!60, fill=black!20, very thick, minimum size=10mm},
container/.style={draw, rectangle, draw=red!60, dashed, inner sep=1em},
]

\node[squarednode]      (state)                      {$s_t$};
\node[squarednode]      (state2)     [right=of state] {$s_{t+1}$};
\node[roundnode]      (action)     [above=of state] {$a_{t}$};
\node[roundnode]        (obs)       [below=of state] {$o_t$};
\node[squarednode]        (state3)       [above=of state2] {$\hat{s}_{t+1}$};
\node[roundnode]        (obs2)       [right=of obs] {$ o_{t+1}$};
\node[container, fit=(state3) (state2)] (or) {};

\draw[-{Latex[length=3mm,width=2mm]}] (obs.north) -- (state.south);
\draw[-{Latex[length=3mm,width=2mm]}] (obs2.north) -- (state2.south);
\draw[-{Latex[length=3mm,width=2mm]}] (action.east) -- (state3.west);
\draw[-{Latex[length=3mm,width=2mm]}] ($(state.east)-(state.south)$) -- ($(state3.south)-(state.east)$);

\end{tikzpicture}

\caption{Forward Model: predicting next state $s_{t+1}$ from $s_t$ and $a_t$. The error is computed between predicted state $\hat{s}_{t+1}$ and the actual next state $s_{t+1}$}
\label{fig:ForwardModel}
\end{figure}
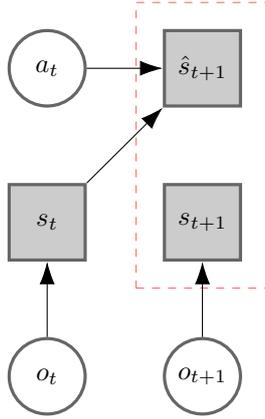
A forward model predicts $s_{t+1}$ from $o_t$ or $s_t$ and $a_t$ (Fig. \ref{fig:ForwardModel}).
In this context, we want to learn the mapping $\phi$ from $o_t$ to $s_t$ using the model that predicts $s_{t+1}$ from $o_t$. Hence the prediction work in two steps, first encoding from $o_t$ to $s_t$ then transition from $s_t$ to $\hat{s}_{t+1}$. We can not compute any error on $s_t$, however at $t+1$ the model can learn from the error between $\hat{s}_{t+1}$ and $s_{t+1}$. The error is back propagated through the transition model and the encoding model. Consequently the methods allows to learn a model $\phi$.

\begin{equation}  
\hat{s}_{t+1} = f(s_t, a_t; \theta_{fwd})
\label{eq:fwd}
\end{equation}

Learning such a model makes it possible to impose structural constraints on the model for state representation learning. For example, the forward model can be constrained to be linear between $s_t$ and $s_{t+1}$, imposing that the system in the learned state space follows simple linear dynamics.

\item{\textbf{Learning an inverse model}}:
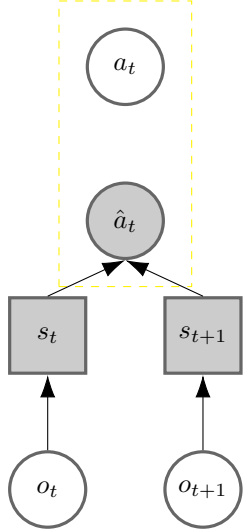
\begin{figure}
\centering
\begin{tikzpicture}[
roundnode/.style={circle, draw=black!60, fill=green!0, very thick, minimum size=10mm},
roundnode2/.style={circle, draw=black!60, fill=black!20, very thick, minimum size=10mm},
squarednode/.style={rectangle, draw=black!60, fill=black!20, very thick, minimum size=10mm},
container/.style={draw, rectangle, draw=yellow, dashed, inner sep=1em},
]

\node[squarednode]      (state)                      {$s_t$};
\node[squarednode]      (state2)     [right=of state] {$s_{t+1}$};

\coordinate (middle) at ($(state.west)!0.5!(state2.east)$);
\node[roundnode2]      (action2)     [above=of middle] {$ \hat{a}_{t}$};
\node[roundnode]      (action)     [above=of action2] {$a_{t}$};
\node[roundnode]        (obs)       [below=of state] {$o_t$};
\node[roundnode]        (obs2)       [right=of obs] {$ o_{t+1}$};
\node[container, fit=(action) (action2)] (or) {};

\draw[-{Latex[length=3mm,width=2mm]}] (obs.north) -- (state.south);
\draw[-{Latex[length=3mm,width=2mm]}] (obs2.north) -- (state2.south);
\draw[-{Latex[length=3mm,width=2mm]}] (state.north) -- (action2.south);
\draw[-{Latex[length=3mm,width=2mm]}] (state2.north) -- (action2.south);


\end{tikzpicture}

\caption{Inverse Model: predicting action $a_t$ from $s_t$ and $s_{t+1}$. The error is computed between predicted action $\hat{a}_t$ and the actual action $a_t$}
\label{fig:InverseModel}
\end{figure}
An inverse model predicts action $a_{t}$ given observations $o_t$ and $o_{t+1}$ or states $s_t$ and $s_{t+1}$.
Like for forward model, the goal here is to learn the mapping $\phi$ from $o_t$ to $s_t$ through two steps, encoding for $o_t$ and $o_{t+1}$ and action prediction. The error is computed for action prediction between $a_t$ and $\hat{a}_t$ and then back-propagated to learn the encoding

\begin{equation}
\hat{a}_t = g(s_t, s_{t+1}; \theta_{inv})
\label{eq:inv}
\end{equation}
Learning such model enforces that the state encodes enough information to recover the action that modified the state (Fig.~\ref{fig:InverseModel}).

\item{\textbf{Using prior knowledge to constrain the state space}}:
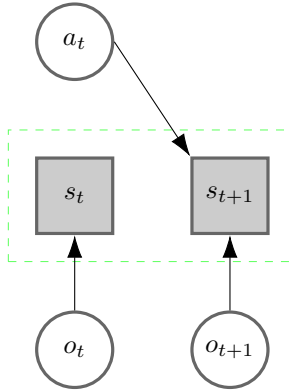
\begin{figure}
\centering
\begin{tikzpicture}[
roundnode/.style={circle, draw=black!60, fill=green!0, very thick, minimum size=10mm},
squarednode/.style={rectangle, draw=black!60, fill=black!20, very thick, minimum size=10mm},
container/.style={draw, rectangle, draw=green!60, dashed, inner sep=1em},
]

\node[squarednode]      (state)                      {$s_t$};
\node[squarednode]      (state2)     [right=of state] {$s_{t+1}$};
\node[roundnode]      (action)     [above=of state] {$a_{t}$};
\node[roundnode]        (obs)       [below=of state] {$o_t$};
\node[roundnode]        (obs2)       [right=of obs] {$ o_{t+1}$};
\node[container, fit=(state) (state2)] (or) {};

\draw[-{Latex[length=3mm,width=2mm]}] (obs.north) -- (state.south);
\draw[-{Latex[length=3mm,width=2mm]}] (obs2.north) -- (state2.south);
\draw[-{Latex[length=3mm,width=2mm]}] (action.east) -- ($(state2.west)-(state.south)$);

\end{tikzpicture}
\caption{Model with prior: The error is computed by applying loss functions on several states}
\label{fig:Priors}
\end{figure}
A last approach is to handle SRL by using specific constraints or prior knowledge about the functioning, dynamics or physics of the world (besides the constraints of forward and inverse models) such as the temporal continuity or the causality principles that generally reflect the interaction of the agent with objects or in the environment \cite{Jonschkowski15}. \textit{Priors} are defined as objective or loss functions $\mathcal{L}$, applied on a set of states $s_{1:n}$ (Fig. \ref{fig:Priors}), to be minimized (or maximized) under specific condition $c$. An example of condition can be enforcing locality or time proximity within the set of states.

\begin{equation}
Loss = \mathcal{L}_{prior}(s_{1:n}; \theta_{\phi} | c)
\label{eq:prior}
\end{equation}

\end{itemize}

All these approaches are detailed in Section \ref{sec:LearningObjectives}.

\subsection{State representation characteristics} 

Besides the general idea that the state representation has the role of encoding essential information (for a given task) while discarding irrelevant aspects of the original data, let us detail what the characteristics of a good state representation are. 

In a reinforcement learning framework, the authors of \cite{Bohmer15} defines a good state representation as a representation that is:
\begin{itemize}
\item Markovian, i.e. it summarizes all the necessary information to be able to choose an action within the policy, by looking only at the current state.
\item Able to represent the true value of the current state well enough for policy improvement.
\item Able to generalize the learned value-function to unseen states with similar futures.
\item Low dimensional for efficient estimation. 
\end{itemize} 

Note that these are some characteristics expected of the state representation, but they cannot be used for learning this representation. Instead, they can later be verified by assessing  the task performance for a controller based on the learned state. Note also that multiple state representations can verify these properties for a given problem and that therefore, there is no unique solution to the state representation learning problem. We detail this problem when discussing the evaluation of the learned state space in Section \ref{sec:evaluation}.

State representation learning can also be linked with the idea of learning disentangled representations that clearly separate the different factors of variation with different semantics. Following \cite{Achille17}, a good representation must be sufficient, as efficient as possible (i.e., easy to work with, e.g., factorizing the data-generating factors), and minimal (from all possible representations, take the most efficient one).

The \textit{minimal} assumption is comparable to the simplicity prior \cite{Jonschkowski15}. It assumes that only a small number of world properties are relevant, and that there exists a low dimensional state representation of a higher level input observation. Related to \textit{Occam’s razor}, this prior favors state representations that exclude irrelevant information to encourage a lower dimensionality.

The \textit{efficiency} aspect of the representation means that there should be no overlapping between dimensions of the learned state features. Unfortunately, independence of features alone may not be enough to assure a good quality of representations and guarantee a disentanglement of factors of variation \cite{Thomas17}.
Higher level abstractions can, however, allow to improve this disentanglement and permit easier generalization and transfer. Cues to disentangle the underlying factors can include spatial and temporal scales, marginal independence of variables, and controllable factors \cite{Thomas17}.

\subsection{State representation learning applications}
\label{sec:srl-applications}
The main interest of SRL is to produce a low dimensional state space in which learning a control policy will be more efficient. Indeed, deep reinforcement learning in the observation space has shown spectacular results in control policy learning \cite{Mnih15,Lillicrap15,Mnih16} but is known to be computationally difficult and requires a large amount of data \cite{Rusu2016}. Separation of representation learning and policy learning is a way to lighten the complete process. 
As described in most of the reviewed papers \cite{Mattner12,Watter15,Hoof16,Munk16,Curran16,Wahlstrom15,Shelhamer17,Oh17}, this approach is used to make reinforcement learning faster in time and/or lighter in computation.

SRL can be particularly relevant with multimodal observations that are produced by several complementary sensors with high dimensionality as is, for example, the case of autonomous vehicles. Low dimensional representations are then key to make an algorithm able to take decisions from hidden factors extracted from these complementary sensors. This is for instance the case of representation learning from different temporal signals in \cite{Duan17,Bohg17}. Audio and images are blended in \cite{Yang17} while RGB and depth are combined in \cite{Duan17}.
SRL can also be used in a transfer learning setting by taking advantage of a state space learned on a given task to rapidly learn a related task. This is for example the case in \cite{Jonschkowski15} where a state space related to a robot position is learned in a given navigation task and reused to quickly learn another navigation task. SRL is also used as pretraining for transfer to other applications afterwards such as reinforcement learning \cite{Munk16,Oh17}.
For concrete examples on SRL application scenarios see Section \ref{sec:EvaluationScenarios}.

Another case where SRL could be useful is in the application of Evolution Strategies (ES) for robot control learning \cite{Stulp13}. Evolution strategies are a family of black box optimization algorithms that do not rely on gradient descent and can be an alternative to RL techniques (such as Q-learning and policy gradients) but are less adapted to high-dimensional problems. Indeed, the convergence time of ES algorithm depends on the dimension of the input: the larger the dimension is, the larger amount of solutions ES has to explore \cite{Stulp13}.
ES optimization methods have shown to be efficient for deep reinforcement learning \cite{Clune17} but they could then take a clear advantage of a lower dimension input to explore faster the parameter space than using raw data \cite{Alvernaz17}.

\section{Learning objectives}
\label{sec:LearningObjectives}
In this section, we review what objectives can be used to learn a relevant state representation. A schema detailing the core elements involved in each model's loss function was introduced in Fig.~\ref{fig:AutoEncoder}~--~\ref{fig:Priors}, which highlights the main approaches to be described here. This section touches upon machine learning tools used in SRL such as auto-encoders or siamese networks. A more detailed description of these is later addressed in Section~\ref{sec:tools}.

\subsection{Reconstructing the observation}
\label{sub:recon}
 
A first idea that can be exploited is the fact that a true state, along with some noise, was used to generate the observation. Under the hypothesis that the noise is not too large, compressing the observation should retain the important information contained in the true state. While this idea is very often exploited with dimensionality reduction algorithms \cite{fodor2002survey} such as Principal Component Analysis (PCA), we focus here on the recent approaches specifically dealing with state representation.

The PCA algorithm is a linear transformation able to compress and decompress observations with minimal reconstruction error. PCA have been exploited to reduce the dimensionality of the state space during learning \cite{Curran16}. By projecting images into a 3- or 4-dimensional space, it is possible to produce a state that is used by a reinforcement learning algorithm and that reduces the convergence time in Super Mario games \cite{Karakovskiy12} and different simulations such as Swimmers or Mountain Car.

Auto-encoders are models that learn to reproduce their input under constraints on their internal representations such as dimensionality constraints (Fig. \ref{fig:AutoEncoder}). Their architecture can therefore be used to learn a particular representation in low dimensions $s_t$ by reconstructing $o_t$.

Simple auto-encoders can be used to learn 2D representation of a real pole from raw images \cite{Mattner12} (see Section \ref{sec:EvaluationScenarios} on evaluation scenarios). After training, the encoding vector from the AE is used to learn a controller to balance the pole. An auto-encoder whose internal representation is constrained to represent a position
that serves as input to a controller is also presented in \cite{Finn15} and \cite{Alvernaz17}. The proposed model learns a state representation from raw pixels of respectively a PR2 robot's hand and an agent in the VizDoom environment.

These models, based on auto-encoders that reconstruct the observation at the same time step, can however learn only if the factors of variations are only linked to the actual state, or if very prominent features exists \cite{Lesort17}. In order to relax this assumption, it is possible to reconstruct observations from other time steps or to use constraints on the evolution of the state (as will be more detailed in Section \ref{sec:forward}) to focus reconstruction on features relevant to the system dynamics.

An example of auto-encoder tuned to take into account the system dynamics is proposed in \cite{Goroshin15} where an AE with a siamese encoder projects sequences of images into a state representation space $\mathcal{S}$ with constraints
on the transition between $s_t$ and $s_{t+1}$
to be linear.
They use observations at several time steps in order to take time into account in the representation, and predict future observations through a single decoder that reconstructs $\hat{o}_{t+1}$. This makes the model able to learn representations that are related to several time steps
and filter out random features of the environment.

The idea of using an auto-encoder to learn a projection into a state space where transitions are assumed to be linear has also been used by \cite{Watter15}. The model presented, ``Embed To Control" (E2C), consists of a deep generative model that learns to generate image trajectories from a linear latent space. 

Extending \cite{Watter15}, the representation with dynamic constraints can be learned on policy at the same time as a reinforcement learning algorithm learns a task \cite{Hoof16}.
They compared different types of auto-encoders to learn visual and tactile state representations and use this representation to learn manipulation task policies for a robot.
Sharing the same idea of embedding dynamic constraints into auto-encoders, Deep Variational Bayes Filters (DVBF) are an extension of Kalman filters which learn to reconstruct the observation based on a nonlinear state space using  variational inference \cite{Karl16}.  The reconstruction from a non linear state space based on a model inspired by a Deep Dynamical Model (DDM) \cite{Wahlstrom15} and E2C \cite{Watter15} is proposed in \cite{Assael15}. It is argued that the model is adapted for better training efficiency and it can learn tasks with complex non-linear dynamics \cite{Assael15}. Their result shows improvements over the PILCO model \cite{Deisenroth11}, which learns a state representation by only minimizing the reconstruction error without constraining the latent space.  

\subsection{Learning a forward model}
\label{sec:forward}

Last subsection reviewed how reconstructing an observation is useful  to learn state representations. Now we will review how temporal dynamics of the system can also help the same purpose.
Therefore we present approaches that rely on learning a \textit{forward} model to learn a state space. The general idea is to force states to efficiently encode the information necessary to predict the next state (Fig.~\ref{fig:ForwardModel}).

As described in Section \ref{sec:formalism}, in the case of the forward models we study here, the model is used as a proxy for learning $s_t$. The model firstly makes a projection from the observation space to the state space to obtain $s_t$ and applies a transition to predict $\hat{s}_{t+1}$.  The error is computed by comparing the estimated next state $\hat{s}_{t+1}$ with the value of $s_{t+1}$ derived from the next observation $o_{t+1}$ at the next time step.

Note that forward models can benefit from the observation reconstruction objective presented in Section \ref{sub:recon}. As an example, the works previously presented in Section \ref{sub:recon} \cite{Goroshin15,Hoof16,Watter15,Assael15,Karl16} belong to the auto-encoder category of models. However, they all predict future observations to learn representations and therefore, they as well belong to the family of forward models.

The method these works use to combine forward models and auto-encoders consists in mapping $o_t$ to $s_t$, and then compute the transition, with the help of $a_t$, to obtain $\hat{s}_{t+1}$. $\hat{s}_{t+1}$ is then remapped onto the pixel space in form of a vector $\hat{o}_{t+1}$. The error is then computed pixel-wise between $\hat{o}_{t+1}$ and $o_{t+1}$.
One common assumption is that the forward model in the  learned state space is linear \cite{Goroshin15, Hoof16}. The transition is then just a linear combination of $s_t$ and $a_t$ as in Eq. \ref{eq:linear}. $W, U$ and $V$ are either fixed or learned parameters \cite{Hoof16}.

\begin{equation}
\hat{s}_{t+1}= W * \hat{s}_t + U * a_t + V
\label{eq:linear}
\end{equation}

In a similar way, the \textit{Embed to Control} model (E2C) based on variational auto-encoders uses Eq.~\ref{eq:linear} to compute the mean $\mu$ of a distribution and learn supplementary parameters for the variance $\sigma$ of the distribution~\cite{Watter15}. Then, $\hat{s}_{t+1}$ is computed with Eq. \ref{eq:linear2}:

\begin{equation}
\hat{s}_{t+1} \sim \mathcal{N}  (\mu=W* \hat{s}_t + U * a_t + V, \sigma)
\label{eq:linear2}
\end{equation}

Using distributions to compute $\hat{s}_{t+1}$ allows to use the KL-divergence to train the forward model. This method is also used in \cite{Karl16} and \cite{Krishnan15}. However, the transition model in \cite{Krishnan15} 
considers the KL-divergence

between $P(s_{t+1})$ and $P(\hat{s}_{t+1})$ and does not use the loss of the reconstruction based on $o_{t+1}$ and $\hat{o}_{t+1}$.

The use of $a_t$ is a common feature in most forward models based approaches.
In fact, as several future states $s_{t+1}$ from a given state are possible, $s_t$ alone does not contain enough information to predict $s_{t+1}$. The only approach that gets rid of the need for actions assumes that the transition from $s_{t-1}$ to $s_{t}$ allows to deduce the transition from $s_t$ to $s_{t+1}$ and uses several past states to predict $s_{t+1}$ \cite{Goroshin15}. Actions are therefore implicit in this approach.

Another use of a forward model, connected to an intrinsic curiosity model (ICM) which helps agents explore and discover the environment out of curiosity when extrinsic rewards are sparse or not present at all, is proposed in \cite{Pathak17}. In this model, an intrinsic reward signal is computed from the forward model's loss function $\mathcal{L}_{fwd}$ ($\hat{f}$ is the forward function learned by the model, $\hat{\phi}$ is the encoding model):

\begin{equation}
\mathcal{L}_{fwd} (\hat{\phi}(o_{t+1}), \hat{f}(\hat{\phi}(o_{t}),a_{t})) = \frac{1}{2} \parallel \hat{f} (\hat{\phi}(o_{t}),a_{t}) − \hat{\phi}(o_{t+1}) \parallel_2 ^2
\end{equation} 
It is argued that there is no incentive in this model for $s_t$ to learn to encode any environmental features that cannot influence or are not influenced by the agent's actions. The learned exploration strategy of the agent is therefore robust to uncontrollable aspects of the environment such as the presence of distractor objects, changes in illumination, or other sources of noise in the environment \cite{Pathak17}.

Forward models are therefore able to learn representations of controllable factors: in order to predict next state, the model must understand the object being controlled. This kind of representation can also be learned through a controllability prior \cite{Jonschkowski17}.

If a robot acts by applying forces, controllable things should be those whose accelerations correlate with the actions of the robot. Accordingly, a loss function can be defined to minimize the covariance between an action dimension $i$ and the accelerations in the state dimension $i$. The following formula from \cite{Jonschkowski17} makes it explicit:

\begin{equation} 
Controllability_i = e^{-cov(a_{t,i},s_{t+1,i})},
\label{equation_Prior_controllability}
\end{equation}

where $cov(a_{t,i},s_{t+1,i})$ is the covariance between the state $s_{t+1, i}$ at dimension $i$ and time $t$ and $a_{t,i}$ (the action at dimension $i$ that led to such state). Note that here the learned state is assumed to represent an acceleration. Related with this prior also is the notion of \textit{empowerment}~\cite{Klyubin05}, defined as an information-theoretic capacity of an agent’s actuation channel to influence its own evolution. The concept of empowerment is related to \textit{accountability} or \textit{agency}, i.e., recognizing when an agent is responsible for originating the change of state in the environment.

\subsection{Learning an inverse model}

The forward model approach can be turned around and, instead of learning to predict next state (given previous state and action), use current and next states to predict the action between them. The inverse model framework is used in SRL by firstly performing a projection of $o_t$ and $o_{t+1}$ onto learned states $s_t$ and $s_{t+1}$, and secondly, by predicting the action $\hat{a}_t$ that would explain the transition of $s_t$ into $s_{t+1}$ (Fig. \ref{fig:InverseModel}). As before, learning this model can impose constraints on the state representation to be able to efficiently predict actions. 

An example using inverse models to learn state representations is the Intrinsic Curiosity Module (ICM) \cite{Pathak17}. It integrates both an inverse and forward model 
 and the authors argument that using an inverse model is a way to bypass the hard problem of predicting original observations (e.g., pixels in images), since actions have much lower dimension.

A different kind of inverse model is used in \cite{Shelhamer17}, where the policy gradient algorithm used to learn a controller is augmented with auxiliary gradients from what is called \textit{self-supervised} tasks. In this case, in lack of external supervision, the prediction error resulting from interactions with the environment acts as a self-supervision. They learned a inverse dynamics model to retrieve from $o_t$ and $o_{t+1}$ the action $a_t$ performed between the two successive time steps.

Note that connections among forward and inverse models are important: inverse models can provide supervision to learn representations that the forward model regularizes by learning to predict $s_{t+1}$~\cite{Agrawal16}. In practice, this is implemented by decomposing the joint loss function as a sum of the inverse model loss plus the forward model loss \cite{Agrawal16}. Conversely, \cite{zhang18} shows in an ablation study that using an inverse model (along with a forward model and an auto-encoder) is the factor that contributes the most to learning a good state representation. 
Another approach including forward and inverse models, as well as a reconstruction of the observation including multimodal inputs is \cite{Duan17}.

\subsection{Using feature adversarial learning}
\label{sec:adversarial}

Adversarial networks \cite{Goodfellow14} can also be used for unsupervised learning of state representations. 
The use of the Generative Adversarial Network (GAN) framework to learn state representations is proposed in \cite{Chen16}. They present a model named InfoGAN that achieves the disentanglement of latent variables on 3D poses of objects. As described in \cite{Chen16}, the goal is to learn a generator distribution $P_G(o)$ that matches the real distribution $P_{data}(o)$. Instead of trying to explicitly assign a probability to every $o$ in the data distribution, GANs learn a generator network $G$ that 
samples from the generator distribution $P_G$ by transforming a noise variable $z \sim P_{noise}(z)$ into a sample $G(z)$. 
The noise variable has two components. A first one, $z_G$, randomly sampled from a Gaussian distribution, and a second one with smaller dimension, $z_U$, sampled from a uniform distribution. The latter is used during training so that the $G(z)$ has a high mutual information with $z_U$. Then, the sample from $z_U$ has a high correlation with $G(z)$ and can thus be considered as a state representation.
This generator is trained by playing against an adversarial discriminator network $D$ that aims at distinguishing between samples from the true distribution $P_{data}$ and the generator distribution $P_G$. The authors succeed to learn states corresponding to object orientations from sequences of images.

Another example of SRL with Generative Adversarial Networks is presented by \cite{Donahue16, Dumoulin16}. BiGAN and ALI are extensions of regular GANs to learn the double mapping from image space to latent space, and from latent space to image space. They allow the learned feature representation to be useful for auxiliary supervised discrimination tasks, and competitive with unsupervised and self-supervised feature learning. The BiGAN has also been experimented in \cite{Shelhamer17} to learn state representations used for reinforcement learning, but lead to lower performance than their own approach (Section \ref{sec:forward}).

\subsection{Exploiting rewards}
As opposed to RL, the use of a reward value in SRL is not compulsory. However, it can be used as supplementary information to help differentiating states and to learn task related representations. Rewards are helpful to disentangle meaningful information from a noisy or distracting one, and to tie the representation to a particular task. However, in a multi-task setting, this approach can also be used to learn a generic state representation that is relevant to different tasks.

A \textit{predictable reward prior} which estimates $\hat{r}_{t+1}$ given a state $s_t$ and an action $a_t$ is implemented in \cite{Munk16} (along with a forward model) to learn a state for reinforcement learning. 
Another approach that exploits reward is presented in~\cite{Oh17}. Besides predicting the reward similarly to~\cite{Munk16}, they also learn to predict the value (discounted sum of future reward) of the next state and exploit this capacity to rely on planning multiple steps for policy learning. The author state that predicting rewards for multiple steps is much easier that predicting observations, while giving the important information for learning a policy.

A dimensionality reduction model called \textit{reward weighted principal component analysis} (rwPCA), as another way of using rewards for state representation was proposed in \cite{Parisi17}. \textit{rwPCA} uses data collected by an RL algorithm and operates a dimensionality reduction strategy which takes reward into account to keep the information into a compressed form. The compressed data is afterwards used to learn a policy.

On the same idea of constructing a task-related representation, \cite{Jonschkowski15} and \cite{Lesort17} use rewards as supplementary information to impose constraints on the state space topology. One of these constraints makes the space more suited to discriminate between states with different rewards. The state space is then particularly adapted to solve a given task. This constraint is called \textit{causality prior} in \cite{Jonschkowski15} and \cite{Lesort17}. It assumes that if we have two different rewards after performing the same action in two different time steps, then the two corresponding states should be differentiated and far away in the representation space (Equation \ref{equation_Prior_Caus}).

\begin{equation}
\mathcal{L}_{Caus}(D,\hat{\phi})=\mathbf{E}[ e^{-\parallel\hat{s}_{t_2}-\hat{s}_{t_1}\parallel^2} \mid a_{t_1}=a_{t_2},r_{t_1+1}\neq r_{t_2+1}] 
\label{equation_Prior_Caus}
\end{equation}

\subsection{Other objective functions}
\label{sub:prior}

In this section, we present other approaches assuming various specific constraints for state representation learning. Following~\cite{Lake16}, the learning process can be constrained by prior knowledge (either initially provided by the designer or acquired via learning) to allow the agent to leverage existing common sense, intuitive physics, physical laws, mental states of others, as well as other abstract regularities such as compositionality and causality.
This kind of a priori knowledge is called \textit{prior} \cite{Bengio12}, \cite{Jonschkowski15}, \cite{Lesort17},\cite{Jonschkowski17}, and is defined through cost functions. These loss functions are applied in the state space in order to impose the required constraints to construct the model projecting the observations in the state space. In the following, $\Delta s_t = s_{t+1}-s_t$ is the difference in between states at times $t$ and $t+1$, and $D$ is a set of observations.

\begin{itemize}

\item  \textbf{Slowness Principle}\\
The slowness principle assumes that interesting features fluctuate slowly and continuously through time and that a radical change inside the environment has low probability \cite{Wiskott02,Kompella11}.

\begin{equation}
\mathcal{L}_{Slowness}(D,\phi)=\mathbb{E}[\parallel\Delta\ s_t\parallel^2] 
\label{equation_Prior_Temporel}
\end{equation}

This assumption can have other naming depending on the unit of $s_t$, e.g., prior of time coherence (time) \cite{Jonschkowski15,Lesort17} or inertia (velocity)  \cite{Jonschkowski17}.

\item \textbf{Variability} \\
The assumption of this prior is that positions of relevant objects vary, and learning state representations should then focus on moving objects \cite{Jonschkowski17}.

\begin{equation}
\mathcal{L}_{Variability}(D,\phi)=\mathbb{E}[ e^{- \parallel s_{t1}- s_{t2}\parallel}] 
\label{equation_Prior_Variation}
\end{equation}

$e^{-distance}$ is used as a  similarity measure that is 1 if the distance among states is 0 and goes to 0 with increasing distance between states. Note that this prior is counter-balancing the slowness prior introduced above as the slowness alone would lead to constant values.

\item \textbf{Proportionality}\\
The proportionality prior introduced in \cite{Jonschkowski15} assumes that for the same action in different states, the reactions to this action will have proportional amplitude or effect. The representation then vary in the same amount for two equal actions in different situations.

\begin{equation}
\mathcal{L}_{Prop}(D,\phi)=\mathbb{E}[(\parallel\Delta s_{t_2}\parallel-\parallel\Delta s_{t_1}\parallel)^2 | a_{t_1}=a_{t_2}] 
\label{equation_Prior_Prop}
\end{equation}

\item \textbf{Repeatability}\\
This prior states that two identical actions applied at similar states should provide similar state variations, not only in magnitude but also in direction \cite{Jonschkowski15}.

\begin{equation}
\mathcal{L}_{Rep}(D,\phi)=\mathbb{E}[e^{-\parallel s_{t_2}-s_{t_1}\parallel^2}\parallel\Delta s_{t_2}-\Delta s_{t_1}\parallel^2 \mid a_{t_1}=a_{t_2}]
\label{equation_Prior_Rep}
\end{equation}

\item \textbf{Dynamic verification}\\
Dynamic verification \cite{Shelhamer17} consists in identifying the corrupted observation $o_{t_c}$ in a history of $K$ observations $o_t$ where $t \in \llbracket 0,K \rrbracket$.  Observations are first encoded into states and the sequence is classified by a learned function $f$ to output the corrupted time step. Negative samples are produced by incorporating observations from a wrong time step into the sequence of images. 
This discriminative approach forces SRL to encode the dynamics in the states.

\item \textbf{Selectivity}\\
States can be learned by using the idea that factors such as objects correspond to `independently controllable' aspects of the world that can be discovered by interacting with the environment \cite{Thomas17}. 
Knowing the dimension $K$ of the state space, the aim is to train K policies $\pi_k$ with $k \in \llbracket 1,K \rrbracket$. The goal is that the policy $\pi_k$ causes a change in $s_t^{(k)}$ only, and not in any other feature. To quantify the change in $s_t^{(k)}$ when actions are taken according to $\pi_k$, the selectivity of a feature $k$ is:

\begin{equation}
\mathcal{L}_{sel}(D,\phi,k)=\mathbb{E} \bigg[ \frac{\parallel s_{t+1}^{(k)}-s_t^{(k)}\parallel}{\sum_{k'} \parallel s_{t+1}^{(k')}-s_{t}^{(k')}\parallel} | s_{t+1} \sim P_{s_t, s_{t+1}}^{a} \bigg]
\label{equation_Prior_Sel}
\end{equation}

where  $P_{s_t, s_{t+1}}^{a}$ is the environment transition distribution from $s_t$ to $s_{t+1}$ under action $a$. The selectivity of $s_{t}^{(k)}$ is maximal when only that single feature changes as a result of some action. 
Maximizing the selectivity improves the disentanglement of controllable factors in order to learn a good state representation.

\end{itemize}


\begin{longtable}{|p{20mm}|p{10mm}|p{10mm}|p{10mm}|p{10mm}|p{10mm}|p{10mm}|}
\caption{Classification of the reviewed SRL models with regards to the learning objectives they implement and the information they use (actions or rewards) 
}
\label{tab:models_characteristics}\\

\hline
\textbf{Model} &
\begin{sideways}\textbf{Actions/Next}\end{sideways} \begin{sideways}\textbf{state constraints}\end{sideways}  &
\begin{sideways}\textbf{Forward model}\end{sideways}  &
\begin{sideways}\textbf{Inverse model}\end{sideways}  &
\begin{sideways}\textbf{Reconstruct}\end{sideways} 
\begin{sideways}\textbf{observation}\end{sideways}  &
\begin{sideways}\textbf{Predicts next }\end{sideways} \begin{sideways}\textbf{observation}\end{sideways}  &
\begin{sideways}\textbf{Uses rewards}\end{sideways}  \\\hline\hline

\endfirsthead 

\hline
\textbf{Model} &
\begin{sideways}\textbf{Actions/Next}\end{sideways} \begin{sideways}\textbf{state constraints}\end{sideways}  &
\begin{sideways}\textbf{Forward model}\end{sideways}  &
\begin{sideways}\textbf{Inverse model}\end{sideways}  &
\begin{sideways}\textbf{Reconstruct}\end{sideways} 
\begin{sideways}\textbf{observation}\end{sideways}  &
\begin{sideways}\textbf{Predicts next }\end{sideways} \begin{sideways}\textbf{observation}\end{sideways}  &
\begin{sideways}\textbf{Uses rewards}\end{sideways}  \\\hline\hline
\endhead



AE \cite{Mattner12} &
no  & 
 no  & 
no & 
\cellcolor{gray} yes & 
 no  & 
no\\\hline 

Priors \cite{Jonschkowski15} &
\cellcolor{gray} yes  & 
 no  & 
no  & 
no & 
no  & 
\cellcolor{gray} yes\\\hline 

PVE \cite{Jonschkowski17} &
\cellcolor{gray} yes  &
 no  &
no  &
no & 
no  &
no\\\hline

E2C \cite{Watter15} &
\cellcolor{gray} yes  &
\cellcolor{gray} yes  &
no  &
\cellcolor{gray} yes & 
\cellcolor{gray} yes  &
no\\\hline

ML-DDPG \cite{Munk16} &
\cellcolor{gray} yes  & 
\cellcolor{gray} yes  & 
no  & 
no & 
no  & 
\cellcolor{gray} yes\\\hline 

VAE/DAE \cite{Hoof16} &
\cellcolor{gray} yes  & 
\cellcolor{gray} yes  & 
no  & 
\cellcolor{gray} yes & 
\cellcolor{gray} yes  & 
no \\\hline 

AE \cite{Finn15} &
\cellcolor{gray} yes  & 
no  & 
no & 
\cellcolor{gray} yes & 
no  & 
no\\\hline 

DVBF \cite{Karl16} &
\cellcolor{gray} yes  & 
\cellcolor{gray} yes  & 
no & 
\cellcolor{gray} yes & 
\cellcolor{gray} yes  & 
no\\\hline 

\cite{Goroshin15} &
\cellcolor{gray} yes  & 
\cellcolor{gray} yes  & 
no & 
\cellcolor{gray} yes & 
\cellcolor{gray} yes  & 
no\\\hline 

ICM \cite{Pathak17} &
\cellcolor{gray} yes  & 
\cellcolor{gray} yes  & 
\cellcolor{gray} yes & 
no & 
no  & 
no\\\hline 

\cite{Shelhamer17} &
\cellcolor{gray} yes  & 
no  & 
\cellcolor{gray} yes & 
no & 
no  & 
no\\\hline 

VPN \cite{Oh17} &
no  & 
\cellcolor{gray} yes  & 
no & 
no & 
no  & 
\cellcolor{gray} yes\\\hline 

DDM \cite{Assael15} &
\cellcolor{gray} yes  & 
\cellcolor{gray} yes  & 
no & 
\cellcolor{gray} yes & 
\cellcolor{gray} yes  & 
no\\\hline 

AE \cite{Wahlstrom15} &
\cellcolor{gray} yes  & 
\cellcolor{gray} yes  & 
no & 
\cellcolor{gray} yes & 
\cellcolor{gray} yes  & 
no\\\hline 

\cite{Thomas17} &
\cellcolor{gray} yes  & 
no  & 
no & 
\cellcolor{gray} yes & 
 no  & 
no\\\hline 

PCA \cite{Curran15} &
no  & 
 no  & 
no & 
\cellcolor{gray} yes & 
 no  & 
no\\\hline 

PCA \cite{Curran16} &
no  & 
 no  & 
no & 
\cellcolor{gray} yes & 
 no  & 
no\\\hline 

rwPCA \cite{Parisi17} &
no  & 
 no  & 
no & 
\cellcolor{gray} yes & 
 no  & 
\cellcolor{gray} yes\\\hline 

\cite{Magrans18} &
\cellcolor{gray} yes  & 
\cellcolor{gray} yes  & 
no & 
no & 
no  & 
\cellcolor{gray} yes\\\hline 
 
InfoGAN \cite{Chen16} &
no  & 
no & 
no & 
\cellcolor{gray} yes & 
no  & 
no\\\hline 

BiGAN \cite{Donahue16} &
no  & 
no & 
no & 
\cellcolor{gray} yes & 
no  & 
no\\\hline 

\cite{Duan17} &
\cellcolor{gray} yes  & 
\cellcolor{gray} yes & 
\cellcolor{gray} yes & 
\cellcolor{gray} yes & 
\cellcolor{gray} yes & 
no\\\hline 

DARLA \cite{Higgins16} &
no  & 
no & 
no & 
\cellcolor{gray} yes & 
no  & 
no\\\hline 

\cite{zhang18} &
\cellcolor{gray} yes  & 
\cellcolor{gray} yes & 
\cellcolor{gray} yes & 
\cellcolor{gray} yes & 
\cellcolor{gray} yes  & 
no \\\hline 

AE \cite{Alvernaz17} &
no  & 
no & 
no & 
\cellcolor{gray} yes & 
no  & 
no \\\hline 

world model \cite{Ha18} &
yes  & 
yes & 
no & 
yes & 
no  & 
no \\\hline 

\end{longtable}

\subsection{Using hybrid objectives}

Reconstruction of data in the observation space, forward models, inverse models, exploitation of rewards, and other objective functions presented in the previous sections are different approaches to tackle the state representation learning challenge. However, these approaches are not incompatible, and models often take advantage of several objective functions at the same time.

For instance, interactively \textit{learning to poke by poking} \cite{Agrawal16} is an example of empirical learning of intuitive physics using $o_t$ and $o_g$ as current and goal images, respectively, in order to predict the poke action. The latter is composed by the location, angle and length of the action that sets the object in the state of goal image $o_g$.
Simulations shows that using the inverse model or jointly the inverse and forward models improve performance at pushing objects and that when the training data available is reduced, 
the joint model outperforms the inverse model with a performance comparable to using a considerably larger amount of data.

\begin{figure}
\centering
\begin{tikzpicture}[
roundnode/.style={circle, draw=black!60, fill=green!0, very thick, minimum size=10mm},
roundnode2/.style={circle, draw=black!60, fill=black!20, very thick, minimum size=10mm},
squarednode/.style={rectangle, draw=black!60, fill=black!20, very thick, minimum size=10mm},
container_Prior/.style={draw, rectangle, draw=green!60, dashed, inner sep=1em},
container_AE/.style={draw, rectangle, draw=blue, dashed, inner sep=1em},
container_fwd/.style={draw, rectangle, draw=red!60, dashed, inner sep=1em},
]

\node[squarednode]      (state)                      {$s_t$};
\node[squarednode]      (state2)     [right=of state] {$s_{t+1}$};
\node[roundnode]      (action)     [above=of state] {$a_{t}$};
\node[roundnode]        (obs)       [below=of state] {$o_t$};
\node[squarednode]        (state3)       [above=of state2] {$\hat{s}_{t+1}$};
\node[roundnode]        (obs2)       [right=of obs] {$ o_{t+1}$};
\node[roundnode2]        (obs3)       [right=of obs2] {$\hat{o}_{t+1}$};
\node[container_fwd, fit=(state3) (state2)] (fwd) {};
\node[container_AE, fit=(obs2) (obs3)] (ae) {};
\node[container_Prior, fit=(state) (state2)] (ae) {};

\draw[-{Latex[length=3mm,width=2mm]}] (obs.north) -- (state.south);
\draw[-{Latex[length=3mm,width=2mm]}] (obs2.north) -- (state2.south);
\draw[-{Latex[length=3mm,width=2mm]}] (action.east) -- (state3.west);
\draw[-{Latex[length=3mm,width=2mm]}] ($(state3.east)+(state.south)$) -- (obs3.north);
\draw[-{Latex[length=3mm,width=2mm]}] ($(state.east)-(state.south)$) -- ($(state3.south)-(state.east)$);

\end{tikzpicture}

\caption{Example of hybrid model (E2C \cite{Watter15}) combining a set of previously described \textit{loss term blocks} of previous figures. Several errors are used to learn the state representation using at the same time a forward model error, a reconstruction error and a constraint on the representation that enforces the transition to be linear.}
\label{fig:AllModels}
\end{figure}
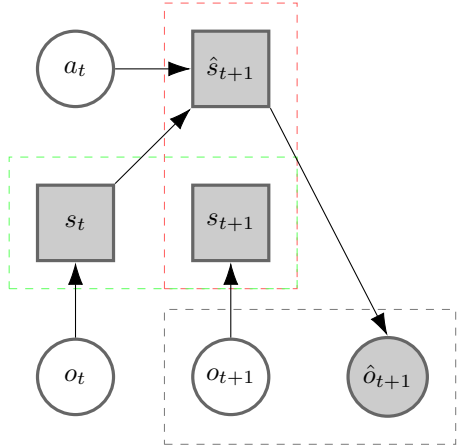

The authors in \cite{Finn15,Goroshin15} use the reconstruction of the observation and the slowness principle in their SRL approach. \cite{Goroshin15,Hoof16,Watter15,Assael15,Karl16,Ha18} combine the reconstruction of observation and forward models. \cite{Jonschkowski15,Lesort17} take advantage of rewards with a causality prior (Eq. \ref{equation_Prior_Caus}) and several  other objective functions such as the slowness principle, proportionality, and repeatability to learn state representations. We illustrate, as an example, the combination of objective functions from \cite{Watter15} in  \figurename~\ref{fig:AllModels}.

Table \ref{tab:models_characteristics} summarizes all the reviewed models by showing, for each one, which proxies or surrogate functions have been used for learning: reconstruction of observation, prediction of the future (forward model) and/or retrieving actions (inverse model), and what kind of information is used: action and/or rewards.

\section{Building blocks of State Representation Learning} 
\label{sec:tools}

In this section, we cover various implementation aspects relevant to state representation learning and its evaluation. We refer to specific surrogate models, loss function specification tools or strategies that help constraining the information bottleneck and generalizing when learning low-dimensional state representations,

\subsection{Learning tools}
We first detail a set of models that through an auxiliary objective function, help learning a state representation. One or several of these learning tools can be integrated in broader SRL approaches as was previously described.

 \subsubsection{Auto-encoders (AE)}

Auto-encoders (AE) are a common tool used to learn state representations that are widely used for dimensionality reduction \cite{Hinton06, Wang12,Wang16}. Their objective is to output a reproduction of the input. Its architecture is composed by an encoder and a decoder. The encoder projects the input to a latent space representation (often in lower dimension than the input), which is re-projected to the output afterwards by the decoder.
In our problem setting, $o_t$ is the input, $s_t$ the latent representation, and $\hat{o}_{t}$ is the output. The dimensionality of the latent representation can be chosen depending on the dimension of the state representation we want to learn and enforcing it in such case.
The AE will then automatically learn a compact representation by minimizing the reconstruction error between input and output.
The usual loss function $\mathcal{L}$ to measure the reconstruction error is the mean squared error (MSE) between input and output, computed pixel-wise. However, it can be any norm.

\begin{equation}
Loss = \mathcal{L}( x,  \hat{x}) 
\label{equation_Squared_error}
\end{equation}

Auto-encoders are used in different SRL settings \cite{Finn15,Mattner12};  PCA can also be considered as a particular case of auto-encoder \cite{Curran16}.

\subsubsection{Denoising auto-encoders (DAE)}

The main issue of auto-encoders is the risk of finding an easy but not satisfying solution to minimize the pixel reconstruction error. This occurs when the decoder reconstructs a kind of \textit{average} looking dataset. To make the training more robust to this kind of mean optimization solution, denoising auto-encoders (DAE) \cite{Vincent08,Vincent10} can be used. This architecture adds noise to the input and makes the ``average" image a more corrupted solution than the original AE. 
The DAE architecture is used in \cite{Hoof16} to learn visual and tactile state representations. The authors compared state representations learned by a DAE and a variational auto-encoder (VAE) by using the learned states in a reinforcement learning setting. They found that, in most cases, DAE state representation models gather less rewards than those with VAE state representations.

\subsubsection{Variational auto-encoders (VAE)}

The SRL literature has also benefited from the variational inference used in variational auto-encoders (VAE) \cite{Kingma13,Rezende14} to learn a mapping from observations to state representations. 
A VAE is an auto-encoder with probabilistic hidden cells: it interprets $\mathcal{S}$ as a set sampled from distribution $P(s_t|o_t)$. 
It then approximates $P(s_t|o_t)$ with a model $q_\theta(s_t|o_t)$ called the approximate posterior or recognition model. $\theta$ represents the parameters of the model, which, for instance, can be a neural network. The VAE also provides a generator which approximates $P(o_t|s_t)$ with a model $p_\phi$. $\phi$ represents the parameters of the generator. Both models $p$ and $q$ are then trained by minimizing the error between $o_t$ and $\hat{o}_t$ and the KL divergence between $q_\theta(s_t|o_t)$ and the normal distribution $\mathcal{N}(\mu=0, 
\sigma=\mathds{I}$) (where $\mu$ is the mean of the distribution, $\sigma$ its covariance matrix and $\mathds{I}$ the identity matrix).
VAE-related models that do not use exactly the original VAE, but variations of it, are \cite{Watter15,Assael15,Krishnan15,Hoof16,Karl16,Higgins16}.  

\subsubsection{Siamese networks}

Siamese networks \cite{Chopra05} consist of two or more identical networks that share their parameters, i.e., have the exact same weights. The objective of the siamese architecture is not to classify input data, but to differentiate between the inputs (\textit{same} versus \textit{different} class or condition, for example). This kind or architecture is useful to impose constraints in the latent space of a neural network. For example it can be used to learn similarity metrics or time dependencies, as it is done in time-contrastive networks \cite{Sermanet17}. 

In the context of SRL, siamese networks can be employed to implement some priors previously presented in Section \ref{sub:prior}. For example, two siamese networks can be used to compute a similarity loss and optimize the slowness principle (or temporality prior) between $s_t$ and $s_{t+1}$ as in \cite{Lesort17}. In \cite{Goroshin15} they use three siamese networks to compute three consecutive states at the same time that are fed into another model that predicts the next state.

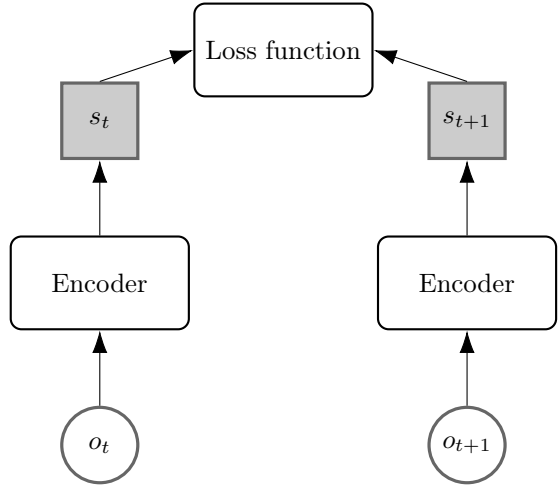
\begin{figure}
\centering
\begin{tikzpicture}[
roundnode/.style={circle, draw=black!60, fill=green!0, very thick, minimum size=10mm},
squarednode/.style={rectangle, draw=black!60, fill=black!20, very thick, minimum size=10mm},
container/.style={draw, rectangle, dashed, inner sep=1em},
box/.style={draw, rectangle, rounded corners, thick, node distance=7em, text width=6em, text centered, minimum height=3.5em},
]


\node[box]      (Encoder)                      {Encoder};
\node[box]      (Encoder2)        [right=of Encoder] {Encoder};
\node[squarednode]      (state)   [above=of Encoder] {$s_t$};
\node[roundnode]      (obs)   [below=of Encoder] {$o_t$};
\node[squarednode]      (state2)   [above=of Encoder2] {$s_{t+1}$};
\node[roundnode]      (obs2)   [below=of Encoder2] {$o_{t+1}$};

\coordinate (middle) at ($(Encoder.west)!0.5!(Encoder2.east)$);
\node[box]      (loss)        [above=of middle] {Loss function};

\draw[-{Latex[length=3mm,width=2mm]}] (obs.north) -- (Encoder.south);
\draw[-{Latex[length=3mm,width=2mm]}] (Encoder.north) -- (state.south);
\draw[-{Latex[length=3mm,width=2mm]}] (obs2.north) -- (Encoder2.south);
\draw[-{Latex[length=3mm,width=2mm]}] (Encoder2.north) -- (state2.south);

\draw[-{Latex[length=3mm,width=2mm]}] (state.north) -- (loss.west);
\draw[-{Latex[length=3mm,width=2mm]}] (state2.north) -- (loss.east);

\end{tikzpicture}

 \caption{Representation of siamese encoders: networks with tied (shared) parameters. In SRL, siamese networks allow to set constraints via loss functions among several states, e.g. in robotic priors. }
 \label{fig:siamois}
\end{figure}

\subsection{Observation/action spaces}
\label{sec:obs_act_spaces}
This section presents a summary of the dimensionality of the observation, state and action spaces, as well as the applications in which the reviewed papers are evaluated (Table~\ref{tab:dimension_table}). The continuity or discreetness of the action space is also shown. These are good proxies to assess the complexity of the problem tackled: the higher the dimensionality of observation and action, as well as the smaller we want the dimension of the state to be, the harder is the task of learning a state representation because much more information will need to be processed and filtered in order to keep only the information that is substantial.

We note also that the reviewed literature  
often presents results with presumably higher dimensionality of learned states than theoretically needed (e.g. using state of dimension 6 for a 2 joints robotic arm). The dimensionality of the state may seem obvious when we are learning a state that should (according to the task) correlate with a clear dimension (position, distance, angle) in the environment. However, deciding the dimensionality of the state space is not always trivial when we are learning more abstract states with no clear dimension associated to it. For instance, visually representing the state associated to an Atari game scene in a complex situation is not as easy to interpret nor assess in comparison to the dimensionality of states associated to a position of an arm, its angle or velocity. Indeed, since the learning objective we reviewed are just proxies to guide state representation learning, they can lead to something different that the ideal and minimal state representation. In particular, increasing the capacity of the model by augmenting the dimensionality of the state above the dimension of the true state can lead to a better optimization of the learning objectives.


\begin{longtable}{|p{20mm}|p{16mm}|p{7mm}|p{15mm}|p{30mm}|p{18mm}|}
\caption{Settings of each approach: characteristics of each environment and state representation dimensions 
}
\label{tab:dimension_table}\\
\hline
\textbf{Reference} &

\textbf{Observa-tion }
\textbf{Dimension}&
\textbf{State}
\textbf{Dimension} &
\textbf{Action}
\textbf{Dimension}  &
\textbf{Environment}  &
\textbf{Data}  \\\hline\hline
\endfirsthead

\hline
\textbf{Reference} &
\textbf{Observa-tion }
\textbf{Dimension}&
\textbf{State}
\textbf{Dimension} &
\textbf{Action}
\textbf{Dimension}  &
\textbf{Environment}  &
\textbf{Data}  \\\hline\hline
\endhead

Priors \cite{Jonschkowski15} &
16*16*3  &
2  &
25 discrete   &
Slot cars, mobile robot localization &
Raw images\\\hline

PVE\cite{Jonschkowski17} &
\cellcolor{gray}Unavailable  &
5 &
Discrete   &
Inverted pendulum, ball in cup, cart-pole &
Raw images\\\hline

E2C \cite{Watter15} &
40*40*3  &
8  &
Discrete   &
Agent with obstacle&
Raw images\\\hline

E2C \cite{Watter15} &
48*48*3  &
8  &
Discrete   &
Inverted pendulum&
Raw images\\\hline

E2C \cite{Watter15} &
80*80*3  &
8  &
Discrete   &
Cart-pole&
Raw images\\\hline

E2C \cite{Watter15} &
128*128*3  &
8  &
Discrete   &
3 link arm&
Raw images\\\hline

\cite{Hoof16} &
20*20*3  &
3  &
Continuous   &
Pendulum swing-up&
Raw images\\\hline

\cite{Hoof16} &
228  &
3  &
Continuous   &
Real-robot manipulation task &
Tactile data\\\hline

ML-DDPG \cite{Munk16} &
18 or 24 &
6  &
2 discrete   &
2 link arm&
Joint position \\\hline

ML-DDPG \cite{Munk16} &
192 or 308   &
96  &
36 discrete   &
Octopus &
Joint position\\\hline

\cite{Finn15} &
240*240*3  &
32  & 
Continuous   &
Robotics manipulation tasks &
Raw images\\\hline

DVBF \cite{Karl16} &
16*16*3  &
3  & 
\cellcolor{gray}Unavailable  &
Pendulum &
Raw images\\\hline

DVBF \cite{Karl16} &
16*16*3  &
2  & 
\cellcolor{gray}Unavailable   &
Bouncing ball &
Raw images\\\hline

DVBF \cite{Karl16} &
16*16*3  &
12  & 
\cellcolor{gray}Unavailable   &
2 bouncing balls &
Raw images\\\hline

\cite{Goroshin15} &
3 frames of 32*32  &
2  & 
2 discrete   &
NORB dataset&
Raw images\\\hline

ICM \cite{Pathak17} & 
42*42*3  & 
3   & 
4 discrete   & 
3D VizDoom navigation game & 
Raw images\\\hline 

ICM \cite{Pathak17} & 
42*42*3  & 
2   & 
14 discrete   & 
Mario Bros & 
Raw images\\\hline 

\cite{Shelhamer17} & 
\cellcolor{gray}Unavailable  & 
\cellcolor{gray} Un.   & 
\cellcolor{gray}Unavailable   & 
Atari & 
Raw images\\\hline 

VPN \cite{Oh17} & 
3*10*10  & 
\cellcolor{gray} Un.   & 
4 discrete   & 
2D navigation & 
Raw images\\\hline 

VPN \cite{Oh17} & 
4*84*84  & 
\cellcolor{gray} Un.  & 
4 discrete    & 
Atari & 
Raw images\\\hline 

\cite{Curran16} & 
\cellcolor{gray}Unavailable & 
4  & 
5 discrete    & 
Mountain car 3D & 
\cellcolor{gray}Unavailable\\\hline 

\cite{Curran16} & 
\cellcolor{gray}Unavailable& 
12  & 
243 discrete    & 
6 link swimmer & 
\cellcolor{gray}Unavailable\\\hline 

 \cite{Higgins17} &
  64*64*3&
 64 &
 8 discrete & 
 Mujoco, Jaco arm  &
 Raw images
 \\\hline 
 
  \cite{Higgins17} &
 84*84*3 &
  64 &
 99 discrete & 
 DeepMind Lab  &
 Raw images
 \\\hline 

rwPCA \cite{Parisi17} & 
21*21*3  & 
Auto.  & 
2 continuous   & 
Picking a coin and putting it on a goal  & 
Raw images \\\hline 

\cite{Parisi17} & 
20  & 
Auto.  & 
2 continuous   & 
Hit a ball with a ping-pong paddle & 
Position of objects \\\hline 

\cite{Magrans18} &
40  &
2   &
Continuous  &
Explore a 3 room simulated 2D map & 
Position of agent\\\hline

\cite{Duan17} &
 240*240*4&
 512  &
4 continuous & 
 Poking cube  &
 Raw images + depth
 \\\hline 
 
 \cite{zhang18} &
 10*10*9&
 256  &
4 discrete & 
 Maze  &
 feature vector
 \\\hline 
 
  \cite{zhang18} &
 \cellcolor{gray} Unavailable &
 200  &
 continuous & 
 Mujoco  &
 joints
 \\\hline 

%


\end{longtable}

\subsection{Evaluating learned state representations}
\label{sec:evaluation}

This section provides a review of validation metrics and embedding quality evaluation techniques used across the literature. These are summarized in Table \ref{tab:metrics}.

\begin{table}[!htbp] 
\caption{Evaluation methods for state representation learning and their respective objective.}
\label{tab:metrics}
\begin{center}
\begin{small}
\begin{tabular}{|p{68mm}|p{68mm}|}
\hline
\textbf{Metric} &
\textbf{ Evaluation objective} \\\hline\hline

Task performance \cite{Jonschkowski15,Jonschkowski17,Munk16,Hoof16,Finn15,Pathak17,Shelhamer17,Oh17,Parisi17,Assael15,Higgins16,zhang18,Alvernaz17} &
Assesses that the information needed to solve a task is contained in $s_t$. The evaluation is done in reinforcement learning algorithms.    \\\hline

Disentanglement metric score \cite{Higgins16} &
Measures the disentanglement of the 
latent factors. It assumes that generative factors are known and interpretable. Used in transfer learning, object recognition.   \\\hline

Distortion \cite{Indyk01} &
Measures the preservation of local and global geometry coherence  in unsupervised representation learning.    \\\hline

NIEQA (Normalization Independent Embedding Quality Assessment) \cite{Zhang12} & 
Measures the local \& global neighborhood embedding quality assessment;  not limited to isometric embeddings. Used in manifold learning. \\\hline 

KNN-MSE \cite{Lesort17} &
Measures the degree of preservation of the same neighbors in between the latent space and the ground truth. 
\\\hline

Supervised regression \cite{Jonschkowski17} &
Evaluates the performance of a supervised regression between the learned states and the ground truth. \\\hline
 
\end{tabular}
\end{small}
\end{center}
\end{table}

The most common way of evaluating the quality of the learned state space is by letting an agent use the states to learn a control task, and thus assessing whether the representation is general enough to be transferable.
This method is for example applied to evaluate the performance of an SRL algorithm using reinforcement learning  
\cite{Jonschkowski15,Jonschkowski17,Munk16,Hoof16,Finn15,Pathak17,Shelhamer17,Oh17,Parisi17,Assael15}. 

However, this approach is often very costly and inefficient in terms of time, computation and data. Also, various state-of-the-art RL algorithms may be applied to learn a policy and may result in very different performances for a given state representation.
The uncertainty inherent to RL therefore makes RL algorithms sufficient but not practical nor appropriate to be a necessary condition to validate a particular state representation.
In consequence, it would be desirable to have an intermediate manner to assess the representation that is independent of the algorithm applied to complete the task and there are, indeed, several more direct ways to assess the learned state space. For example, visual assessment of the representation's quality can be done using a Nearest-Neighbors approach as in \cite{Sermanet17, Pinto16}. The idea is to look at the nearest neighbors in the learned state space, and for each neighbor, retrieve their corresponding observation. Visual inspection can then reveal if these two observations indeed correspond to nearest neighbors in the ground truth state space $\tilde{s}$ we intend to learn. 

While the nearest neighbor coherence can be assessed visually, KNN-MSE is a quantitative metric derived from this qualitative information \cite{Lesort17}. Using the ground truth state value for every observation, KNN-MSE measures the distance between the value of an observation and the value of the nearest neighbor observations retrieved in the learned state space. A low distance means that a neighbor in the ground truth is still a neighbor in the learned representation, and thus, local coherence is conserved.

For an observation $o$, KNN-MSE is computed using its associated learned state $s=\phi(o)$ as follows: 
\begin{equation}\label{eq:knn_mse_crit}
\textrm{KNN-MSE}(s)=\frac{1}{k}\sum_{s' \in KNN(s,k) } || \tilde{s} - \tilde{s}' ||^2
\end{equation}
where $\textrm{KNN}(s,k)$ returns the $k$ nearest neighbors of $s$ (chosen with the Euclidean distance) in the learned state space $\mathcal{S}$, $\tilde{s}$ is the ground truth associated to $s$, and $\tilde{s}'$ is the ground truth associated to $s'$.

One of the characteristics that a good representation should possess is to produce a disentangled representation of variation factors. The evaluation of these characteristics can be done using the selectivity prior (see Section \ref{sub:prior} and Eq. \ref{equation_Prior_Sel}) from \cite{Thomas17}. This prior cares about the independence among variations of the representation under each action. However, it is applicable mainly if actions are known to be independent. 

Another way to quantitatively compare the degree of disentanglement reached by a model is using the disentanglement metric score \cite{Higgins16}. It assumes that the data is generated by a process in which the generative factors are known, interpretable, and that some are conditionally independent. 
In order to measure the disentanglement, it uses a simple low-capacity and low VC-dimension linear classifier's accuracy (reported as \textit{disentanglement metric score}). The classifier’s goal is to predict 
the generative factor that was kept fixed for a given 
difference between pairs of representations from the same latent factor.

Other metrics from the area of manifold learning can be used, such as distortion \cite{Indyk01} and NIEQA \cite{Zhang12}; both share the same principle as two quantitative measures of the global quality of a representation: the representation space should, as much as possible, be an undistorted version of the original space.

Distortion \cite{Indyk01} gives insight of the quality of a representation by measuring how the local and global geometry coherence of the representation changes with respect to the ground truth. It was designed in the \textit{embeddings} context as a natural and versatile paradigm for solving problems over metric spaces.

NIEQA (Normalization Independent Embedding Quality Assessment) \cite{Zhang12} is a more complex evaluation than distortion that measures the local geometry quality and the global topology quality of a representation. NIEQA local part checks if the representation is locally equivalent to an Euclidean subspace that preserves the structure of local neighborhoods. NIEQA objectives are aligned with KNN-MSE \cite{Lesort17}, as a measure to assess the quality of the representation, especially locally. The global NIEQA measure is also based on the idea of preserving original structure in the representation space, but instead of looking at the neighbors, it samples ``representative” points in the whole state space. Then, it considers the preservation of the geodesic distance between those points in the state space.

One last mechanism to assess SRL methods is to use supervised learning to learn a regression from the learned representation to its ground-truth \cite{Jonschkowski17}. The training and test sets are separated into two datasets to evaluate if the regression can generalize to unseen states. The assumption is that this regression measures how well meaningful features are encoded in the state. A good generalization would show a good encoding. 

\subsection{Evaluation scenarios}
\label{sec:EvaluationScenarios}
Datasets used to validate state representation learning include varied, but mainly simulated, environments because they are easier to reproduce and generate. Unlike in image recognition challenges where MNIST digits or ImageNet datasets prevail, in state representation learning, a varied set of regular video games or visuomotor tasks in robotics can be found as a test suite for robotics control. Examples of simulated environments include, among others:
\begin{itemize}
\item Pendulum (Inverted or classical): The goal is to represent the state of the pendulum \cite{Watter15,Jonschkowski17,Hoof16,Mattner12}. The pendulum starts in a random position, and the objective is to swing it up so it stays upright (there is no specified reward threshold at which the task is considered solved).

\item Cart-Pole: consists of an inverted pendulum attached to a cart which moves along a frictionless track; the system is controlled applying +1 or -1 force to the cart, and a reward of +1 is provided for every time step that the pole remains upright. The episode ends when the pole is more than 15 degrees from vertical, or the cart moves more than 2.4 units from the center\footnote{\url{https://github.com/openai/gym/wiki/Leaderboard\#pendulum-v0}} (\cite{Watter15}, \cite{Jonschkowski17}).

\item Atari games \cite{Bellemare13}: mostly low (2D) dimensional simulated environments with different agents and goals. In these games, states can be represented through different variables (time in achieving a task, amount of bonus, keeping alive, etc.) \cite{Shelhamer17,Oh17}. 

\item More advanced test games include \textit{VizDoom}, where the levels passed, reward accumulated and exploration levels are used as evaluation metrics \cite{Pathak17,Alvernaz17}. Likewise, Mario Benchmark \cite{Karakovskiy12} is a platform designed for reinforcement learning based on the ``Super Mario Bros" video game. This test suite is for example experimented in \cite{Curran16,Pathak17}. 

\item Other evaluation benchmarks tested in the reviewed works in this survey include simulated octopus arms \cite{Engel06,Munk16}, labyrinths \cite{Thomas17}, navigation grids \cite{Magrans18,Oh17}, driving cars \cite{Jonschkowski15}, or \textit{mountain car} scenarios \cite{Curran16}.
Another example is the \textit{bouncing ball}, where the goal is to learn a representation of one bouncing ball position in 2D (x,y) \cite{Karl16}.

\item In the robotics domain we can find benchmarks on robot manipulation skills \cite{Finn15,Hoof16} such as Baxter pushing a button \cite{Lesort17}, grasping \cite{Finn15}, stabilizing \cite{Hoof16}, poking objects \cite{Agrawal16,Duan17} or balancing a real pendulum \cite{Mattner12}. Nevertheless, some approaches achieve to learn in real environment scenarios, for instance, with mobile robots that explore an arena \cite{Jonschkowski15}.

\end{itemize}

Many of the latter simulated scenarios are part of Universe and OpenAI Gym 
\cite{Brockman16} or DeepMind Labs \cite{Beattie16}. These benchmarking tasks used in the most prominent state representation learning literature are summarized in Table \ref{tab:dimension_table}.

\section{Discussion and future trends}
 
In this section, we first discuss the implications of SRL for autonomous agents and the assessment, comparison and reproducibility of the representation learned. Finally, we explore the consequences of SRL on the interpretability of machine learning algorithms.

\subsection{SRL models for autonomous agents}

SRL methods provide unsupervised tools for autonomous agents to learn representations about the environment without extra annotations. They need, however, that the agent gathers data to learn.
Therefore, the role of environment exploration is an important dimension to investigate in SRL. 
If the space is not sufficiently explored by the agent, acquisition of varied observations and exposure to actions that lead to optimal performance can be hindered \cite{Pathak17}. 

One way to incorporate exploration in SRL is to integrate curiosity or intrinsic motivations \cite{Oudeyer07} in the algorithm that collects data. The overall idea of this approach is to complement the extrinsic reward by an intrinsic reward that favors states where SRL makes the most progress. This is done for example in the Intrinsic Curiosity Module (ICM) \cite{Pathak17} by defining an intrinsic reward linked to the forward model error which encourages exploration. This approach is improved in \cite{Magrans18} by balancing this exploratory behavior with an homeostatic drive to also favor actions that lead to familiar state-action pairs. The reverse question of how the learned state space can influence the performance of intrinsic motivation approaches \cite{Pere18} is also relevant. The automatic exploration, designed to maximize the quality of a learned state representation, is a field to be further explored in order to build high quality representations. 

Another approach to gather enough relevant data could be to perform data augmentation by adding data from simulation; however, the problem is to make the model benefit from simulation data for real life applications, a problem that is known as the \textit{reality gap} \cite{Mouret13}. Nevertheless, using both kinds of data was shown to improve results in particular applications \cite{Bousmalis17}. An interesting research direction is therefore to study how to exploit simulation data to improve SRL for real world applications.

Another problem to ultimately perform SRL autonomously (i.e., without manual parameter tuning) is the choice of the state representation dimension, which is made empirically in most reviewed approaches. 
The challenge of deciding the dimensionality automatically can be related to a bias-variance trade-off, as the dimensionality of the representation constrains the capacity of the model. Indeed, increasing the states dimension augments the capacity of the model, which, as a result, will be better at reducing the training error, but also leads to overfitting.  As discussed in Section \ref{sec:obs_act_spaces}, learning criteria can be better optimized by models with large capacity, and thus, an automatic process is prone to over estimate the dimension needed. 

To avoid choosing manually the state dimension, it is possible to choose automatically a number of features from a larger set such that they have a certain variance and are orthogonal \cite{Parisi17}. 
It can be done by using PCA to produce a set of features in which the most significant ones are selected with respect to the chosen variance. PCA can also be modified to select reward related components \cite{Parisi17}. 
Although the variance has to be fixed a priori, the authors claim that this is usually easier than choosing the state dimension. Extending this technique to other state representation approach could be an interesting research direction.

\subsection{Assessment, comparison and reproducibility in SRL}

The assessment challenge of SRL is two-sided. First, there is no easy nor certified way for validating a learned representation. Secondly, the lack of common evaluation frameworks makes a fair comparison between approaches difficult.

As mentioned in Section \ref{sec:evaluation}, the most objective method to evaluate the quality of representations is to check if the state representation learned can be used by an RL algorithm to solve a task more efficiently.  
However, this assessment is uncertain and unstable given the stochasticity of reinforcement learning algorithms \cite{Henderson17}. Moreover, it is not obvious which RL algorithm is the best choice, and thus, several should be used in the comparison. In practice, a large amount of policy evaluation runs is therefore required in order to provide a robust assessment, which is possible in simulation but is seldom applicable on real robots, given the robots fragility and experimentation time involved. In this case, it is therefore interesting to use several of the other measures presented in section \ref{sec:evaluation}, that only give partial information on the state representation quality, but are possible to apply for a comparison with a ground truth state. 

Comparing approaches from published results is also particularly hard because of the high variability of the environments and data used in the different approaches (as illustrated in Table~\ref{tab:dimension_table}). This points to the need of an evaluation framework incorporating several tasks and several evaluation metrics similar to the ones proposed for reinforcement learning such as the \textit{DeepMind Control Suite} \cite{Tassa18}. Reproducibility guidelines with proper experimental techniques and reporting procedures, as pointed in \cite{Henderson17} for RL, should also be defined for SRL. In the mean time, as there is not yet an ideal method for state representation assessment, 
researchers should at least make public their simulation environment (with possible ground truth), and use simulation settings from other approaches to provide fairer comparisons and facilitate the method reproducibility. Furthermore, we strongly encourage authors to entirely describe their experiments, in particular, report their data and models' characteristics.

\subsection{Providing interpretable systems}

In 2018, European Union regulations on algorithmic decision-making include a ``right to explanation", ``right to opt-out" and ``non discrimination" of models \cite{Goodman16}. Artificial intelligence research is thus granted with an opportunity to further provide meaningful explanations to why algorithms work the way they do. The interpretability of results in machine learning is however a challenging problem that needs proper definition \cite{Lipton16}. In any case, monitoring the degree to which AI systems show the same thinking flows as humans is invaluable and crucial; not only to explain how human cognition works, but also to help AI make better and more fair decisions \cite{Lake16}.

We define interpretability in the SRL context as the capacity for a human to be able to link a variation in the representation to a variation in the environment, and be able to know why the representation was sensitive to this variation. 
As SRL is designed to be able to give this level of interpretability, it could help improving the understanding we have about learning algorithms' output.
 Indeed, the higher the dimension, the less interpretable the result is for humans.  Therefore, the dimensionality reduction induced by SRL, coupled with the link to the control and possible disentanglement of variation factors, could be highly beneficial to improve our understanding capacity of the decisions made by algorithms using this state representation. 

\section{Conclusion}
State Representation Learning algorithms are designed to find a way to compress high-dimensional observation data into a low and meaningful dimensional space for controlled systems. These models only require the agent's observations, its performed actions and, optionally, the reward of the associated task. 
 
This work aims at presenting an accessible guide to learn about SRL approaches, the existing tools for evaluation and the common simulation settings used as benchmark. We presented the learning objectives of the state of the art approaches on SRL and their resemblances and differences. We discussed afterwards the use of SRL for autonomous agents, the difficulties for comparing existing approaches and the interpretability of results.

A general advice when building SRL models would be to integrate as many learning objectives as possible, depending on the available data. As an example, one could use a reconstruction objective for linking the state space to the observations, combined with a predictive objective (forward model) to capture dynamics, and a reward-based objective to apprehend the effects of actions performed. More general priors could also be added to force the state space to be coherent and understandable for humans. While many models integrate several of these objectives, no proposed model currently includes all of them together.

As SRL is designed to automatically learn representations from a set of unlabeled observations, it could be used in future work to learn from evolving environments and could be a step towards continual or lifelong learning. Another area to explore in the future is the integration of exploration strategies for data collection specifically designed to be able to improve the state representation learned. 

\section{Acknowledgements}
This research is funded by the DREAM project under the European Union's Horizon 2020 research and innovation program under grant agreement No 640891. We acknowledge Olivier Sigaud, Antonin Raffin, Cynthia Liem and other colleagues for insightful and detailed feedback.

\bibliographystyle{apalike}
\bibliography{references}

\newpage
\appendix

\end{document}